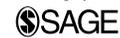
ASSOCIATION FOR
PSYCHOLOGICAL SCIENCE

# Facial Width-to-Height Ratio Does Not Predict Self-Reported Behavioral Tendencies

## Michal Kosinski
Graduate School of Business, Stanford University

## Abstract
A growing number of studies have linked facial width-to-height ratio (fWHR) with various antisocial or violent behavioral tendencies. However, those studies have predominantly been laboratory based and low powered. This work reexamined the links between fWHR and behavioral tendencies in a large sample of 137,163 participants. Behavioral tendencies were measured using 55 well-established psychometric scales, including self-report scales measuring intelligence, domains and facets of the five-factor model of personality, impulsiveness, sense of fairness, sensational interests, self-monitoring, impression management, and satisfaction with life. The findings revealed that fWHR is not substantially linked with any of these self-reported measures of behavioral tendencies, calling into question whether the links between fWHR and behavior generalize beyond the small samples and specific experimental settings that have been used in past fWHR research.

## Keywords
facial width-to-height ratio, fWHR, five-factor model, personality, intelligence, satisfaction with life, impulsiveness, sense of fairness, impression management, self-monitoring, sensational interests, open data

A growing number of studies have linked facial width-to-height ratio (fWHR; Weston, Friday, & Liò, 2007) with various antisocial or violent behavioral tendencies in men, but not in women. For example, Hehman, Leitner, Deegan, and Gaertner (2013) found that fWHR correlated positively with explicit (but not implicit) racial prejudice in a sample of 70 males, $r = .21$, 95% confidence interval (CI) = [−.03, .43], $p = .04$, one-tailed. Stirrat and Perrett (2012) found that broader-faced males were more self-sacrificing when competing with other groups, $n = 15$, $r = .53$, 95% CI = [.02, .82], $p = .04$, but were less self-sacrificing when competing within their own group, $n = 17$, $r = −.50$, 95% CI = [−.79, −.03], $p = .04$. In another study, Stirrat and Perrett (2010) found that broader-faced men (but not women) were more likely to exploit the trust of others in a trust game, $n = 36$, $r = −.40$, 95% CI = [−.64, −.08], $p = .015$, but were not less trusting themselves. Haselhuhn, Wong, Ormiston, Inesi, and Galinsky (2014) reported a positive correlation between facial width and negotiation success among men, $n = 23$ dyads, $\beta = 0.43$, $p = .04$.

Broader-faced men, but not women, have also been shown to be more likely to cheat when reporting dice rolls, $n = 146$, $t(144) = −1.97$, $p = .05$ (Geniole, Keyes, Carré, & McCormick, 2014); to deceive in a negotiation game, $n = 51$, $b = 7.17$, $p = .04$ (Haselhuhn & Wong, 2012); and to adopt an aggressive strategy in a computer game, $n = 37$, $F(2, 34) = 3.6$, $p = .04$ (Carré & McCormick, 2008). Studies have also shown that fWHR is positively correlated with penalty minutes incurred by male hockey players, $n = 21$, $r = .54$, 95% CI = [.14, .79], $p = .01$ (Carré & McCormick, 2008), and with male and female alpha status in capuchin monkeys, $n = 34$, $b = 7.09$, $p = .03$ (Lefevre et al., 2014).

Unfortunately, fWHR research suffers from three major limitations. First, the mechanism underlying the observed links between fWHR and behavior remains unknown. It

**Corresponding Author:**
Michal Kosinski, Stanford University, Graduate School of Business, 665 Knight Way, Stanford, CA 94305
E-mail: michalk@stanford.edu

has been proposed that the relationship between fWHR and aggressive and antisocial behavioral tendencies is mediated by androgens (Carré & McCormick, 2008), but this theory remains controversial (Bird et al., 2016; Whitehouse et al., 2015). Second, the fWHR literature consists of predominantly underpowered studies based on very small samples. Consider the main results of the 10 landmark fWHR studies introduced in the previous paragraph. Half of them were based on samples smaller than 25, and the average sample size was 40. Only 3 of the 10 were significant at a $p$ level below .04. To some extent, this limitation could be explained by the time and financial costs of conducting laboratory experiments and photographing participants' faces. However, the scarcity of publications reporting null results, combined with the disproportionate number of reported $p$ values just below .05, suggests that fWHR research might not be free from the file-drawer and researcher degrees-of-freedom problems (Gelman & Loken, 2014). Third, there is little empirical evidence for links between fWHR and real-life behavioral tendencies. If the links between fWHR and antisocial and violent behaviors generalize beyond the samples and specific experimental settings used in fWHR research, it is reasonable to expect that fWHR should correlate with well-established scales measuring behavioral tendencies—such as trust, sympathy, altruism, cooperation, impulsiveness, or an interest in violence. To date, however, fWHR research has not produced strong evidence for such links; male fWHR has been reported to correlate only weakly with some of the Psychopathic Personality Inventory–Revised scales (Anderl et al., 2016; Geniole et al., 2014) and with Anderson's Personal Sense of Power scale (Haselhuhn & Wong, 2012).

In the study reported here, I tackled the latter two limitations by conducting a large-scale analysis of the links between fWHR and a battery of well-established self-report measures that are typically used for measuring behavioral tendencies.

## Study 1: fWHR and the Five-Factor Model of Personality

Study 1 employed a sample of 1,692 participants to explore the relationship between fWHR and the five-factor model of personality, including the traits of openness, conscientiousness, extraversion, agreeableness, and neuroticism. The five-factor model has previously been shown to be a good predictor of behaviors typically linked with fWHR, such as criminal acts, antisocial behavior, suicide attempts, substance abuse, behavioral problems in adolescent boys, and prejudice (Ozer & Benet-Martínez, 2006; Sibley & Duckitt, 2008). Additionally, as part of this study, I developed and tested a computerized approach to computing fWHR that was subsequently applied to a much larger sample in Study 2.

## Method

**Sample.** The data for this study were taken from the myPersonality.org data set (Kosinski, Matz, Gosling, Popov, & Stillwell, 2015). MyPersonality.org was a Facebook app that offered its users a range of psychometric tests and feedback on their scores. MyPersonality.org users could opt in to donate their scores and Facebook profile data, including their profile pictures, to be used in research. The data set obtained from myPersonality.org included 2,092,439 profile pictures of 815,884 American, British, and Canadian Facebook users (a given user could have more than one more profile picture). In each profile picture, the location of the face, the outlines of its features, and the orientation of the head were identified using the Face++ computer vision software (Megvii Inc., https://www.faceplusplus.com). Figure 1 illustrates the facial landmarks and pitch, roll, and yaw parameters automatically detected by the Face++ software. The results of the Face++ analysis were used to select a subset of 2,597 images that each contained a single, fully visible face (i.e., no facial landmarks were missing) that was looking directly at the camera (i.e., yaw, pitch, and roll parameters were lower than 1°) and was characterized by a distance of at least 50 pixels between the landmarks marking the center of the eyes.

Next, a hypothesis-blind research assistant reviewed these images and removed those in which the visual quality (contrast, focus, and lighting) was low, the facial outline was obscured by hair or clothing, the facial expression was not neutral, the person was not facing the camera directly, the gender of the participant (in the research assistant's judgment) was inconsistent with the gender reported on his or her Facebook profile, or the face clearly did not belong to the participant (a few participants used an image of a popular celebrity as their profile picture). The resulting sample contained 1,703 facial images of 1,692 participants (58% females); participants' median age (obtained from the Facebook profiles) was 28 years (interquartile range = 26–34, range = 20–71).

**Estimating fWHR.** Carré and McCormick's (2008) methodology was used to manually estimate the fWHR of each image. Two hypothesis-blind research assistants independently measured the distance between the cheekbones (the widest central part of the face), as well as the distance between the philtrum and the midbrow, and their estimates were then averaged. I computed fWHR by dividing the distance between the cheekbones by the distance between the philtrum and the midbrow. Across participants, the average fWHR was 1.73 ($SD$ = 0.16) for females and 1.83 ($SD$ = 0.17) for males. Interrater agreement was comparable with the agreement achieved in previous studies (e.g., Haselhuhn & Wong, 2012). It was high for for the fWHR estimates, $r$ = .86, 95% CI = [.85, .87], $p <$

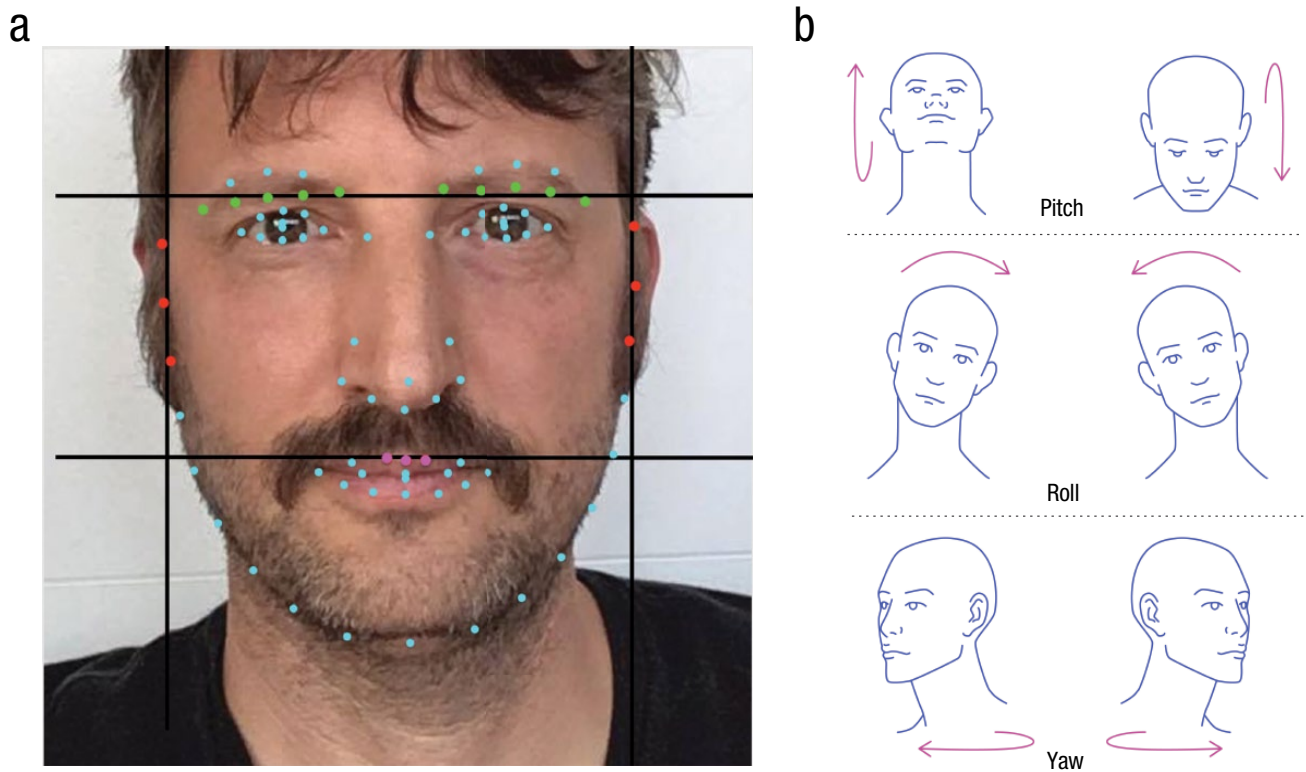

**Fig. 1.** Graphic illustration of the results produced by the Face++ software. The colored dots in (a) show the locations of the 83 facial landmarks (red for cheekbones, pink for upper lip, green for midbrow, and blue for the remaining landmarks). The facial width-to-height ratio was estimated by dividing the distance between the vertical lines marking the cheekbones by the distance between the horizontal lines marking the midbrow and upper lip. The schematics in (b) illustrate the pitch, roll, and yaw parameters describing the head's orientation relative to the camera.

.001 (and for both width and height, $r = .99$, 95% CI = [.99, .99], $p < .001$).

Additionally, fWHR was estimated by using the facial landmarks produced by the Face++ software, by dividing the euclidean distance between the vertical lines crossing the landmarks marking the cheekbones by the distance between the horizontal lines crossing the upper lip and the midbrow (see Fig. 1). The mean fWHR estimate obtained using this computerized approach was 1.65 ($SD = 0.15$) for females and 1.77 ($SD = 0.14$) for males. Thus, the manual and computerized estimates differed systematically by about 0.06. The correlation between the computerized and manual fWHR estimates, $r = .86$, 95% CI = [.85, .87], $p < .001$, was comparable with the interrater agreement of the research assistants.

If more than one facial image was available for a given participant, the fWHR estimates for that participant were averaged.

***Psychometric measures.*** Participants' scores for the five-factor model of personality were measured using the 100-item International Personality Item Pool (IPIP) questionnaire (Goldberg et al., 2006), a widely used instrument that is based on the five-factor model of personality and measures the traits of openness, conscientiousness, extraversion, agreeableness, and neuroticism (see Table 1 for the scales' reliabilities).

## Results

The results, presented in Table 1, revealed that the manual estimates of fWHR were significantly, but weakly, correlated with only women's extraversion, $r = -.081$ (see Table 1 for the 95% CIs for all $r$ values), $p = .032$. This correlation, however, ceased to be significant when the analysis controlled for age, $r = -.071$, $p = .058$ (age correlated positively with fWHR and negatively with extraversion).

The analysis based on computerized fWHR estimates yielded only one weak correlation (see Table 1): with men's conscientiousness, $r = .090$, $p = .005$. This correlation was even weaker when the analysis controlled for age, $r = .060$, $p = .043$ (age correlated positively with both fWHR and conscientiousness).

The patterns of correlation with personality were similar for the manual and the computerized fWHR estimates, $r = .90$, $p < .001$. This finding provided additional evidence for the validity of the computerized approach to estimating fWHR.

**Table 1.** Correlations of Facial Width-to-Height Ratio (fWHR) With Self-Reported Psychological Traits and Age

| Psychometric scale | α | Males | | | Females | | |
|---|---|---|---|---|---|---|---|
| | | *r* | 95% CI | *p* | *r* | 95% CI | *p* |
| Study 1 (manual fWHR estimates) | | | | | | | |
| Five-factor model: domains (987 males, 705 females) | | | | | | | |
| Openness | .84 | −.021 | [−.083, .041] | .510 | −.045 | [−.118, .029] | .237 |
| Conscientiousness | .92 | .059 | [−.003, .121] | .062 | .017 | [−.057, .091] | .650 |
| Extraversion | .93 | −.038 | [−.100, .025] | .239 | **−.081** | [−.154, −.007] | .032 |
| Agreeableness | .88 | .060 | [−.002, .122] | .059 | −.016 | [−.090, .058] | .674 |
| Neuroticism | .93 | .002 | [−.060, .065] | .942 | .041 | [−.033, .115] | .276 |
| Study 1 (computerized fWHR estimates) | | | | | | | |
| Five-factor model: domains (987 males, 705 females) | | | | | | | |
| Openness | .84 | −.019 | [−.081, .043] | .548 | −.056 | [−.129, .018] | .136 |
| Conscientiousness | .92 | **.090** | [.028, .152] | .005 | −.001 | [−.075, .073] | .979 |
| Extraversion | .93 | −.014 | [−.076, .048] | .659 | −.073 | [−.146, .001] | .052 |
| Agreeableness | .88 | .043 | [−.020, .105] | .179 | −.021 | [−.094, .053] | .585 |
| Neuroticism | .93 | −.042 | [−.104, .021] | .189 | .035 | [−.039, .108] | .358 |
| Study 2 (computerized fWHR estimates) | | | | | | | |
| Five-factor model: domains (55,458 males, 81,027 females) | | | | | | | |
| Openness | .84 | **−.042** | [−.050, −.034] | < .001 | **−.054** | [−.061, −.047] | < .001 |
| Conscientiousness | .92 | **.063** | [.055, .071] | < .001 | **.075** | [.069, .082] | < .001 |
| Extraversion | .93 | **.010** | [.002, .019] | .015 | **−.020** | [−.027, −.013] | < .001 |
| Agreeableness | .88 | **.036** | [.028, .045] | < .001 | **.073** | [.066, .080] | < .001 |
| Neuroticism | .93 | **−.030** | [−.038, −.022] | < .001 | **−.034** | [−.040, −.027] | < .001 |
| Five-factor model: facets (223 males, 165 females) | | | | | | | |
| Trust | .90 | .038 | [−.093, .169] | .568 | −.088 | [−.237, .066] | .262 |
| Morality | .79 | .014 | [−.118, .145] | .839 | −.073 | [−.223, .081] | .354 |
| Altruism | .85 | .015 | [−.116, .146] | .820 | −.056 | [−.207, .097] | .472 |
| Cooperation | .71 | .000 | [−.132, .131] | .998 | −.049 | [−.200, .105] | .532 |
| Modesty | .79 | −.115 | [−.242, .017] | .088 | −.073 | [− .224, .080] | .349 |
| Sympathy | .81 | −.043 | [−.173, .089] | .524 | .014 | [−.139, .166] | .858 |
| Self-efficacy | .84 | .035 | [−.097, .166] | .604 | −.062 | [−.213, .092] | .429 |
| Orderliness | .84 | −.008 | [−.139, .124] | .909 | −.142 | [−.288, .011] | .069 |
| Dutifulness | .77 | .011 | [−.120, .143] | .865 | −.038 | [−.190, .116] | .629 |
| Achievement-striving | .84 | −.028 | [−.159, .104] | .678 | −.038 | [−.189, .116] | .632 |
| Self-discipline | .89 | −.035 | [−.166, .096] | .599 | −.023 | [−.175, .130] | .767 |
| Cautiousness | .83 | −.020 | [−.151, .112] | .765 | .063 | [−.091, .214] | .420 |
| Friendliness | .89 | .115 | [−.017, .242] | .087 | .004 | [−.149, .156] | .963 |
| Gregariousness | .89 | .077 | [−.055, .206] | .254 | .019 | [−.134, .171] | .808 |
| Assertiveness | .86 | .111 | [−.021, .239] | .098 | −.070 | [−.221, .083] | .369 |
| Activity level | .73 | −.049 | [−.179, .083] | .464 | −.001 | [−.154, .152] | .993 |
| Excitement-seeking | .84 | .009 | [−.123, .140] | .896 | −.095 | [−.244, .059] | .227 |
| Cheerfulness | .86 | −.030 | [−.161, .102] | .658 | −.009 | [−.162, .144] | .906 |
| Anxiety | .87 | −.096 | [−.224, .036] | .155 | .093 | [−.060, .243] | .234 |
| Anger | .92 | −.130 | [−.257, .002] | .053 | .053 | [−.100, .204] | .497 |
| Depression | .91 | −.077 | [−.206, .055] | .255 | .076 | [−.078, .226] | .334 |
| Self-consciousness | .85 | −.127 | [−.255, .004] | .057 | .104 | [−.049, .253] | .182 |
| Immoderation | .78 | .016 | [−.115, .147] | .810 | .061 | [−.093, .212] | .438 |
| Vulnerability | .88 | −.120 | [−.248, .011] | .073 | .133 | [−.020, .280] | .088 |
| Imagination | .84 | −.059 | [−.189, .073] | .379 | −.001 | [−.153, .152] | .995 |

**Table 1.**  *(continued)*

| Psychometric scale | α | Males | | | Females | | |
|---|---|---|---|---|---|---|---|
| | | r | 95% CI | p | r | 95% CI | p |
| Artistic interests | .82 | .032 | [−.100, .163] | .636 | .029 | [−.124, .181] | .708 |
| Emotionality | .77 | −.116 | [−.244, .015] | .084 | .028 | [−.126, .180] | .724 |
| Adventurousness | .83 | **.133** | [.002, .260] | .047 | −.141 | [−.287, .012] | .071 |
| Intellect | .83 | .037 | [−.094, .168] | .578 | −.087 | [−.237, .067] | .266 |
| Liberalism | .80 | −.066 | [−.195, .066] | .330 | −.044 | [−.196, .109] | .572 |
| Barratt Impulsiveness Scale: total score (347 males, 427 females) | .85 | −.035 | [−.140, .071] | .516 | .036 | [−.059, .131] | .454 |
| Barratt Impulsiveness Scale: domains (251 males, 301 females) | | | | | | | |
| Attentional impulsiveness | .72 | −.041 | [−.164, .084] | .521 | .070 | [−.076, .150] | .519 |
| Motor impulsiveness | .68 | −.001 | [−.125, .122] | .982 | −.002 | [−.115, .111] | .976 |
| Nonplanning impulsiveness | .73 | −.050 | [−.172, .075] | .435 | .022 | [−.092, .134] | .707 |
| Barratt Impulsiveness Scale: facets (251 males, 301 females) | | | | | | | |
| Attention | .71 | −.071 | [−.193, .054] | .264 | .016 | [−.097, .129] | .781 |
| Cognitive instability | .52 | .024 | [−.101, .147] | .710 | .051 | [−.062, .163] | .375 |
| Motor | .73 | .005 | [−.119, .128] | .941 | −.024 | [−.137, .089] | .673 |
| Perseverance | .25 | −.013 | [−.137, .111] | .833 | .053 | [−.060, .165] | .360 |
| Self-control | .72 | −.051 | [−.174, .073] | .420 | −.004 | [−.117, .109] | .941 |
| Cognitive complexity | .47 | −.028 | [−.151, .096] | .659 | .050 | [−.063, .162] | .384 |
| Sensational Interests Questionnaire (2,415 males, 3,792 females) | | | | | | | |
| Militarism scale | .79 | .038 | [−.002, .078] | .062 | .011 | [−.020, .043] | .484 |
| Violent-Occult scale | .66 | **−.047** | [−.087, −.007] | .020 | **−.123** | [−.155, −.092] | < .001 |
| Intellectual Recreation scale | .57 | −.027 | [−.067, .012] | .177 | **−.040** | [−.072, −.009] | .013 |
| Occult Credulousness scale | .75 | **−.064** | [−.104, −.024] | .002 | **−.087** | [−.118, −.055] | < .001 |
| Wholesome Activities scale | .69 | .015 | [−.025, .055] | .462 | .027 | [−.005, .059] | .096 |
| Self-Monitoring Scale (772 males, 1,049 females) | | | | | | | |
| Total score | .69 | −.010 | [−.080, .061] | .791 | .016 | [−.045, .076] | .613 |
| Satisfaction With Life Scale (1,680 males, 2,531 females) | | | | | | | |
| Total score | .86 | **.074** | [.026, .121] | .002 | **.085** | [.046, .124] | < .001 |
| Proxy for Raven's matrices (284 males, 292 females) | | | | | | | |
| Total score | | −.020 | [−.136, .096] | .734 | −.003 | [−.118, .112] | .960 |
| Rust's scales (654 males, 862 females) | | | | | | | |
| Sense-of-Fairness scale | .75 | .013 | [−.064, .089] | .747 | .015 | [−.052, .082] | .657 |
| Impression Management scale | .61 | .021 | [−.055, .098] | .584 | .037 | [−.030, .104] | .276 |
| Age (55,568 males, 81,126 females) | | **.069** | [.061, .077] | < .001 | **.129** | [.122, .136] | < .001 |

Note: Correlations significant at the $p < .05$ level are highlighted in bold. The $p$ values in this table were not corrected for multiple comparisons, and the correlation coefficients were not corrected for attenuation due to measurement error (of fWHR or the psychometric scales). The proxy for Raven's matrices was administered as a computerized adaptive test and scored using item response theory, so Cronbach's α is not available for this measure. See the text for the sources of the psychometric measures used. Scatterplots illustrating the relationships between fWHR and these personality variables are available in the Supplemental Material. CI = confidence interval.

## Study 2: fWHR, Intelligence, and Other Personality Traits

In Study 2, I employed a much larger sample ($N = 170,241$) to explore the potential links between fWHR and all 55 psychometric scales available in the myPersonality.org database for which fWHR could be estimated for at least 100 individuals. These scales measured the five domains and 30 facets of the five-factor model of personality, impulsiveness, sensational interests, sense of fairness, impression management, self-monitoring, satisfaction with life, and intelligence. Many of these traits have previously been shown to strongly predict behavioral tendencies that, given the results of past research, should be associated with fWHR. Intelligence, for example, has been shown to be inversely linked with aggression (Giancola & Zeichner, 1994; Huesmann, Eron, & Yarmel, 1987) and cooperativeness (Jones, 2008). Both impulsiveness and sensational interests have been linked with aggression, violent tendencies, and criminal behavior (Egan & Campbell, 2009; Stanford et al., 2009).

### Method

**Sample.** The same data set as the one in Study 1 was used, but the inclusion criteria for the faces were relaxed. Images were included if the distance between the eyes was at least 20 pixels, yaw and pitch were below 2°, and roll was below 9° (note that head roll does not affect the fWHR). The resulting sample contained 170,241 facial images of 137,163 participants (70% females); participants' median age was 27 years (interquartile range = 24–33, range = 17–82).

**Estimating fWHR.** Given the large number of facial images, I used only the computerized approach from Study 1 to estimate fWHR. The mean fWHR was 1.81 ($SD$ = 0.16) for females and 1.87 ($SD$ = 0.16) for males. The estimated fWHRs were averaged if more than one facial image was available for a given participant. For participants who had two facial images available ($n$ = 27,293), the correlation between the two fWHR estimates was significant, $r$ = .60, 95% CI = [.59, .61], $p$ < .001, and this correlation increased for a subset of these participants ($n$ = 219) whose facial images were of the highest resolution (distance between the eyes > 40 pixels), $r$ = .76, 95% CI = [.70, .81], $p$ < .001. This reveals the degree to which computerized fWHR estimates are affected by the resolution of the images.

**Psychometric measures.** In Study 2, I analyzed data from the 100-item IPIP questionnaire, but also included data from a range of other well-established psychometric measures: the 336-item IPIP Personality Facets questionnaire (Goldberg et al., 2006), the Barratt Impulsiveness Scale (30 items; Patton, Stanford, & Barratt, 1995), the Satisfaction With Life Scale (5 items; Diener, Emmons, Larsen, & Griffin, 1985), Rust's Sense-of-Fairness and Impression Management scales (36 items; Rust & Golombok, 1989), the Self-Monitoring Scale (25 items; Snyder, 1974), the Sensational Interests Questionnaire (28 items; Egan et al., 1999), and myPersonality's 20-item proxy for Raven's Standard Progressive Matrices (Raven, 2008). These scales have been previously shown to be of high reliability and high external validity (Kosinski et al., 2015; Kosinski, Stillwell, & Graepel, 2013; see Table 1 for the scales' reliabilities).

### Results

The results, presented in Table 1, revealed that fWHR did not substantially correlate with any of the 55 scales. As in Study 1, fWHR correlated only weakly with scores for the five personality domains ($|r|$ ranging from .010 to .075; see Table 1 for $r$ values and 95% CIs); however, those correlations were significant because of the very large number of cases ($n$ = 136,485). The correlations were even weaker when the analysis controlled for age (the decrease in $|r|$ ranged from .01 to .03). Taken together, participants' scores for the five personality domains explained a negligible amount of variance in fWHR: $R^2$ = .010 for females and $R^2$ = .005 for males.

Other significant correlations were equally weak: Estimated fWHR was positively correlated with satisfaction with life for both men, $r$ = .074, $p$ < .01, and women, $r$ = .085, $p$ < .001. Results for the Sensational Interests Questionnaire showed that fWHR correlated negatively with the Violent-Occult and Occult Credulousness scales for both men and women, and negatively with the Intellectual Recreation scale for women ($r$s ranging from −.040 to −.123). Among the personality facets, the only significant correlation was the one for men's adventurousness, $r$ = .133, $p$ = .047; this relationship ceased to be significant, however, when the analysis controlled for multiple comparisons (using Holm correction). Intelligence, impulsiveness, self-monitoring, sense of fairness, and impression management were not significantly correlated with fWHR. Controlling for age and race (estimated using the Face++ algorithm) did not substantially change the results (see Table 1 for correlations between age and fWHR). Similarly, there were no substantial nonmonotonic relationships between fWHR and any of the psychological traits. (See the Supplemental Material available online for scatterplots illustrating the relationship between fWHR and the psychometric scores.)

The correlations observed in this study remained weak even when corrected for attenuation caused by the limited reliability of the psychometric questionnaires and fWHR estimates. For example, one of the strongest correlations, between fWHR and women's conscientiousness, increased from .075 to .101 when corrected for attenuation.[1]

### Discussion

Overall, the few weak relationships found in this study do not seem to support the links between fWHR and antisocial or violent behavioral tendencies among men observed in past fWHR research. First, in Study 2, all the significant correlations were stronger for women than for men, which contradicts the notion that fWHR is more strongly related to male behavioral tendencies.

Second, none of the 130 correlations provided any support for the links between fWHR and antisocial or violent behavioral tendencies. For example, fWHR was positively correlated with agreeableness among both men and women; broader-faced people reported themselves to be more (not less, as the fWHR literature suggests) prosocial, sympathetic, trusting, and cooperative. Also, both male and female fWHR correlated negatively with the Violent-Occult scale of the Sensational Interests Questionnaire; in other words, broader-faced people reported less interest in drug use, weapons, piercing, and tattoos. Moreover, broader-faced people did not score significantly higher on any of the traits positively related to antisocial and aggressive behavioral tendencies, including the personality facets of excitement seeking and anger, impulsiveness, and militarism (i.e., interest in paramilitary groups, the armed forces, bodybuilding, martial arts, and survivalism). Additionally, broader-faced people did not score significantly lower on any of the traits negatively related to antisocial and aggressive behavioral tendencies, such as intelligence and the personality facets of morality, altruism, and cooperation.

The results do not necessarily indicate that psychological traits are not linked with facial morphology. The observed correlations were all weak, but also highly consistent in direction across genders. With the exception of extraversion, whenever the correlations were significant for both genders, they had the same sign for men and women. These weak correlations, however, are also inconsistent with the results of previous fWHR studies, calling into question whether the reported links between fWHR and behavior generalize beyond the small samples and specific experimental settings used in fWHR research.

## Action Editor

Brent W. Roberts served as action editor for this article.

## Author Contributions

M. Kosinski is the sole author of this article and is responsible for its content.

## Acknowledgments

The author would like to thank Emily Reit, Alisa Yu, Vivian Xiao, Andrea Freund, and Poruz Khambatta for their critical reading of this manuscript; Isabelle Abraham for proofreading the manuscript; and Mariia Vorobiova for designing the figures. Finally, I would like to thank the creators of Face++ for allowing me to use their software free of charge.

## Declaration of Conflicting Interests

The author declared that he had no conflicts of interest with respect to his authorship or the publication of this article.

## Supplemental Material

Additional supporting information can be found at http://journals.sagepub.com/doi/suppl/10.1177/0956797617716929

## Open Practices

All data are available via the Open Science Framework and can be accessed at https://osf.io/vq49b/. The complete Open Practices Disclosure for this article can be found at http://journals.sagepub.com/doi/suppl/10.1177/0956797617716929. This article has received the badge for Open Data. More information about the Open Practices badges can be found at https://www.psychologicalscience.org/publications/badges.

## Note

1. The estimated reliabilities were .92 (Cronbach's α) for conscientiousness and .60 (interrater agreement) for fWHR.